\documentclass{article}

\usepackage{arxiv}
\usepackage[utf8]{inputenc} 
\usepackage[T1]{fontenc}    
\usepackage{hyperref}       
\usepackage{url}            
\usepackage{booktabs}       
\usepackage{amsfonts}       
\usepackage{nicefrac}       
\usepackage{microtype}      
\usepackage{lipsum}
\usepackage{graphicx}
\usepackage{multicol}

\title{Three Branches: Detecting Actions with Richer Features}

\makeatletter
\newcommand{\printfnsymbol}[1]{%
  \textsuperscript{\@fnsymbol{#1}}%
}
\makeatother

\author{
  Jin Xia\thanks{Both authors contributed equally to this work. }\\
  Machine Vision and Intelligence Group\\
  Shanghai Jiao Tong University\\
  \And
  Jiajun Tang\printfnsymbol{1}\\
  Machine Vision and Intelligence Group\\
  Shanghai Jiao Tong University\\
  \And
  Cewu Lu\thanks{Cewu Lu is corresponding author, lucewu@sjtu.edu.cn } \\
  Machine Vision and Intelligence Group\\
  Shanghai Jiao Tong University\\
}

\begin{document}
\maketitle
\begin{abstract}
We present our three branch solutions for International Challenge on Activity Recognition at CVPR 2019. This model seeks to fuse richer information of global video clip, short human attention and long-term human activity into a unified model. We have participated in two tasks: Task A, the Kinetics challenge and Task B, spatio-temporal action localization challenge. For Kinetics, we  achieve 21.59\% error rate. For the AVA challenge, our final model obtains 32.49\% mAP on the test sets, which outperforms all submissions to the AVA challenge at CVPR 2018 for more than 10\% mAP. As the future work, we will introduce human activity knowledge \cite{HAKE} which is a new dataset including key information of human activity. 

\end{abstract}


\section{Introduction}
Video understanding has made considerable progresses in recent years thanks to the evolution of deep learning and available large-scale datasets. C3D \cite{c3d} uses 3D convolutional nerual networks to extract spatio-temporal features from videos, showing that 3D CNN is a good descriptor for action recognition tasks. Based on 3D CNN, P3D \cite{p3d}, S3D \cite{s3d} and I3D \cite{i3d} are proposed to better model the spatio-temporal features. Recently, SlowFast network \cite{slowfast} applies two pathways to capture visual and motion information respectively in the videos and achieves state-of-the-art results on several action recognition datasets. Inspired by self-attention mechanism \cite{self_attention} first proposed in machine translation task, Non-local Network \cite{non-local}, Action Transformer \cite{action-transformer} and Long-Term Feature Bank(LFB) \cite{lfb} apply scaled dot-product attention to model various kinds of interaction for action recognition. Among them, LFB takes long range temporal relations into account and also achieves state-of-the-art performance on multiple tasks.

Large-scale video datasets have made significant contribution to video understanding. Kinetics-400, Kinetics-600, Kinetics-700 \cite{K400, K600} and Moments in Time \cite{moments} are organized for the basic video action classification task. Kinetics-700 is one of the biggest dataset which contains approximately 650, 000 video clips and covers 700 human action classes. ActivityNet \cite{activity-net}, Charades \cite{charades} and AVA \cite{ava} involves some more complex tasks such as temporal proposal, spatio-temporal localization, etc. AVA, a video dataset of spatio-temporally localized atomic visual actions, contains 430 15-minute video clips. Spatio-temporal labels are provided for one frame per second, with every person annotated with a bounding box and multiple actions, resulting in 1.58M action labels in total. 

In this work, we aim to improve state-of-the-art action models and provide better features for AVA action detection task. To this end, a three branches model is proposed. Our insight is that richer feature would largely help the action detection. Three important features are designed. First, the global feature of the wole video clip is extracted. Second, we believe human region should be attended in short-term. Third, the long-term feature is also important for describe an action. Finally, by fusing those three branches together, our model gets richer feature. 

We pre-train the feature extractor on Kinetics-700 and achieve 21.59\% error rate in Task A. For task B, our final model achieves 32.49\% mAP on the AVA dataset, which outperforms all submissions to the AVA challenge at CVPR 2018 for more than 10\% mAP.

\section{Input Representations}
The objective of the AVA task is to recognize each person's actions every second in the video. To benefit from the well developed deep learning models such as object detection, 3D convolution model and Transformer model, our approach takes following components as input. Noting that all the input can be obtained from the video, we describe them in details for clarity. 

\paragraph{Key Frame} For the spatio-temporal action localization task, we have to find the spatial locations for each person at each second. However, detecting each person in every frame could be time consuming, thus, we choose the middle frame in each second as the key frame. Human detection is performed only on the key frames.

\paragraph{Short Video Clip} Since it could be too memory consuming to feed the entire video into GPUs, we cut videos to 5-second short clips around those chosen key frames. During training and testing stage, frames from these short clips will be taken as input instead of the entire video.  

\paragraph{Long-Term Feature} Using only short clips, the method could be poorly performed due to the limited temporal information. To enrich the temporal reception fields, we extract features of each person in each time step using 3D convolutional models. For each prediction, not only the corresponding short clip but also long range person features are provided.

\section{The Proposed Approach}

\begin{figure}
  \centering
  \includegraphics[width=1.\textwidth]{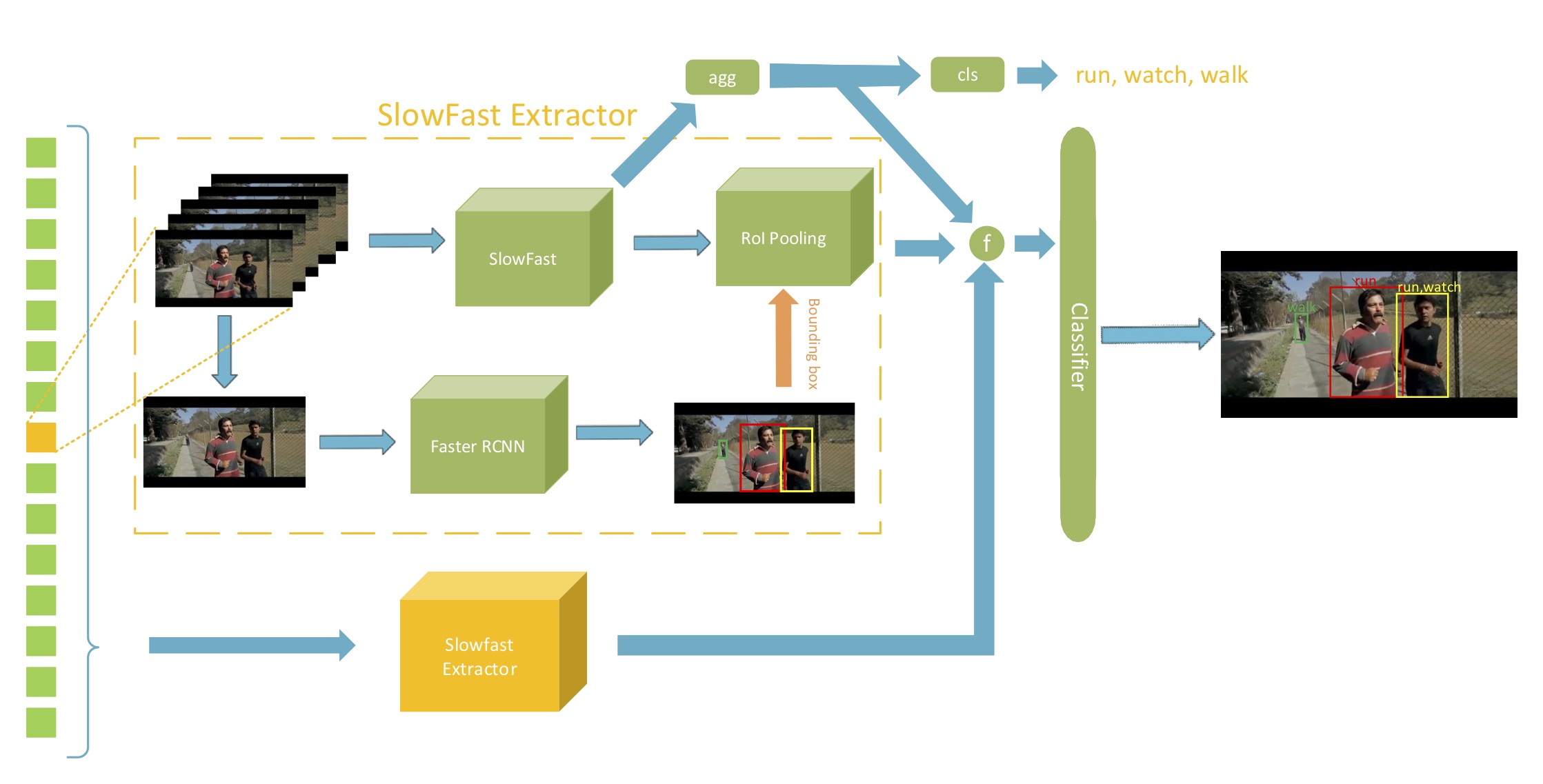}
  \caption{Our method uses SlowFast network as the video feature extractor. Left squares are the video clips. On the bottom branch, a fixed SlowFast network is used to extract features from neighborhood clips. On the middle branch, SlowFast network extracts the video features for the current clip. Human features are cropped using human bounding boxes. On the upper branch, we use a classifier to predict all actions happened in the scenes. The person features, the LFB features and the global features are concatenated and passed to the final classifier.}
  \label{fig:fig1}
\end{figure}

Figure.~\ref{fig:fig1} describes our pipeline for the AVA task. Our final result is fused by three branches. First, SlowFast is used as the feature extractor to extract global clip features for all actions happening in that clip. The global clip features are fed to help final task, by contributing its global view. Second, Faster-RCNN model is used to detect the human boxes. We use 3D Region of Interest (RoI) pooling to crop short-term person features from the clip feature based on human detection box. Third, The short-term person feature and long term feature are then passed to LFB blocks. We fuse short term person feature, long term feature and global scene feature at last and pass it to the final classifier.

\paragraph{Human Detector} The human detector is used to localize actors in clips. One could use any object detection model. We use Faster-RCNN \cite{faster-rcnn} with a
ResNeXt-101-FPN backbone as our human detector. Since we use ground-truth human boxes directly in training phrase, the human detector is only used during testing.  

\paragraph{Backbone} We use SlowFast 3D convolution model because of its high performance. We take 64 contiguous frames in the short clips as the input to the video model. There are two pathways in the model, namely slow pathway and fast pathway. The input of the slow pathway has 8 frames and the input of the fast pathway has 32 frames. We train SlowFast network on Kinetics-600 and fine-tune it on Kinetics-700 for Kinetics challenge. For AVA, we set the spatial stride of $\textrm{res}_5$ to 1 and use a dilation of 2 for its filters following \cite{slowfast}. We extract 3D RoI features at the output feature maps of Slowfast backbone. The 3D RoI Pooling is implemented by replicating 2D RoI Pooling along the temporal axis. For the RoI features, we operate average pooling temporally and max pooling spatially. We find the dilation filters in $\textrm{res}_5$ quite beneficial and a probable reason is that  the expanded reception field helps to merge more global features. 

\paragraph{Long-Term Feature Operation} Temporal dependency could be useful for action recognition. We adapt LFB blocks to capture temporal relations between seconds. In details, we select 305 person features (5 persons per second, 61 seconds in total) around the center clips as long term features. Features in the current clip are considered as short term features. We pre-process these two kinds of input using dropout and dimension reduction. We find these two operations remit the overfitting phenomenons. Apart from the LFB blocks, we tried to use Transformer-style block \cite{self_attention} in which there is a Feed-Forward Network and multi-head Transformers. We didn't find that the Transformer Blocks and the multi-head Transformers gave a better performance. We choose to use two LFB blocks because more than two blocks returns similar results in the experiments. To fuse the output of LFB blocks and the ROI pooled person features, one could use either sum operation, concatenate operation or other fusing methods. We find that concatenate operation have better performance than sum operation.  

\paragraph{Global Loss} Instance-specific RoI feature has concentrated information for each person, it will, however, limit the model to focus on local areas. We observe that more global information could be helpful for action classification. For example, if there is a person playing the musical instrument, there is probably another person listening to music. With this in mind, we design a loss that is aware of all the actions in the scene. We append a per-class Sigmoid-based classifier to predict all the actions that take place in the scene. Noting this loss is not directly related to the task's objective, but it helps capture the interactions between the single person and its global environment. Furthermore, we aggregate the global feature to the RoI feature for the final predictions. Although we use dilated filter in $\textrm{res}_5$ and Non-Local block in backbone model to expand receptive fields, we find the global loss still beneficial. 

\paragraph{Training Loss} The loss mentioned in the baseline method is the per-class Sigmoid loss because AVA is a multi-label task. We modify the training loss in the following directions based on the nature of action classification. Although AVA is a multi-label task, one person could have one and only one pose action. For these pose classes, we use Softmax-based loss instead of Sigmoid-based loss. We also observe that the human action is unbalanced in real life. For example, actions such as talking could happen very frequently, while there could be fewer action instances such as swimming, climbing, etc. Taking the unbalance nature of action class into account, we adopt focal loss \cite{focal_loss} in our method.

\section{Experiments}

\begin{table}
 \caption{Test result on Kinetics-700}
  \centering
  \begin{tabular}{llll}
    \toprule
    Method     & Modalities & Pretrain &  Avg Error Rate (\%)  \\
    \midrule
    Ours     & RGB & Kinetics-600 & \textbf{21.59}  \\
    \bottomrule
  \end{tabular}
  \label{tab:table1}
\end{table}

\begin{table}
 \caption{Test results on AVA}
  \centering
  \begin{tabular}{llll}
    \toprule
    Method     & Modalities & Pretrain & test mAP (\%) \\
    \midrule
    YH Technologies \cite{yh} & RGB + Flow & Kinetics-400 & 19.60 \\
    A better baseline \cite{better_baseline} & RGB & Kinetics-600 & 21.91 \\
    SlowFast & RGB & Kinetics-600 & 27.10 \\
    LFB & RGB & Kinetics-400 &  27.20 \\
    Action Transformer & RGB & Kinetics-400 & 24.93 \\
    Tsinghua/Megvii (winner 2018) & RGB + Flow & Kinetics-400 & 21.08 \\
    Ours     & RGB & Kinetics-700 & \textbf{32.49}  \\
    \bottomrule
  \end{tabular}
  \label{tab:table2}
\end{table}

In this section, we present details for our experiments. For the person detection, we initialize the detector with ImageNet pre-train model. We train the detector with COCO keypoint dataset \cite{coco} using only human class and then fine-tune the human detector with AVA dataset. 

For the SlowFast backbone, ResNet-101 is used as its backbone. We first pre-train it with Kinetics-600 dataset and then fine-tune it with Kinetics-700 dataset. We train the model with a batch size 512. Batch normalization statistics are computed on a single GPU. Half consine learning rate scheduler is used in our experiments, where the learning rate at each iteration is defined formally as $0.5 lr_0\cdot\left[\cos (\frac{iter}{iter_{max}})+1\right] $. Here we use $lr_0$ to denote the initial learning rate, and $iter$ and $iter_{max}$ are respectively current and maximum iteration number. Momentum is set as 0.9 and weight decay is $10^{-4}$. A dropout layer with dropout rate 0.5 is used before the final fully connected layer. The input videos are 8 frames for the slow pathway and 32 frames for the fast pathway. All frames are sampled from a video randomly cropped from raw videos in both temporal and spatial dimensions. Scale jittering is used as mentioned in \cite{slowfast}. For pretraining with Kinetics-600, we set the initial learning rate $lr_0$ as 0.4 and train it for 180k iteration totally. Linear warm-up is used in the first 4k iteration. For fine-tuning with Kinetics-700, we set $lr_0$ to 0.1 and $iter_{max}$ to 60k. The learning rate gradually warms up in the early 3k iteration. Following this recipe, our SlowFast networks achieves 21.59\% average of top-1 and top-5 error rate  on Kinetics-700 test set, as shown in Table.~\ref{tab:table1}.

After pre-training SlowFast networks on Kinetics dataset, we then fine-tune them on our AVA task. We train our model using synchronous SGD with a minibatch size of 64 clips, and the batch normalizaion layers are frozen. We train the model for 26k iterations, with a learning rate of 0.016, which is decreased by a factor of 10 at iteration 16k, 21k. We use a weight decay of $10^{-7}$ and momentum of 0.9. We scale the short side of the video to 256 pixels. We use ground-truth human box for training and detected human box with confidence score more than 0.8 for testing. For data augmentation, we perform random flipping, random box jittering and color jittering. Note that we fine-tune two models with AVA dataset here. We first fine-tune the simple SlowFast networks to extract features for the long-term feature banks. Then we fine-tune the whole networks which invoke long-term features and global features. During both fine-tuning processes we initialize from weights pre-trained on Kinetics, but for different part of our models. For the final test, we use test augmentation including multi-scale and horizontal flipping augmentation. We scale the short side of video to [244, 256, 320] during testing. Table.~\ref{tab:table2} displays results of our method and other methods proposed before the challenge in 2019.

\section{Conclusion and Future Work}

We have presented our action models for Kinetics-700 task and AVA task at CVPR 2019.  It's especially encouraging to see that solutions for the AVA task has moved up for more than 10\% mAP from 2018 to 2019. The large scale datasets and the new proposed methods have contributed a lot to the progress of video understanding. However, the precision of action localization is still relatively low especially for the rare actions. The training cost for video understanding is still expensive. As the futher work, on the one hand, a more computing efficient model is needed for researches and industries. Key point feature extracted by well developed model \cite{alpha_pose} can be a for action recognition. We will go along the way to improve deep learning framework. On the other hand, data-driven has its limit, we also seek new direction of combining human activity knowledge \cite{HAKE}.

\bibliographystyle{unsrt}  


\bibliography{template}





\end{document}